\pdfoutput=1
\documentclass[11pt,a4paper]{article}
\usepackage[OT6,T1]{fontenc}
\usepackage[utf8]{inputenc}
\usepackage{CJKutf8}
\usepackage{devanagari}
\usepackage[english]{babel}
\usepackage[hyperref]{acl2020}
\usepackage{times}
\usepackage{latexsym}
\usepackage{multirow}
\usepackage{booktabs}
\usepackage{tipa}

\newcommand{\armenian}{\fontencoding{OT6}\fontfamily{cmr}\selectfont}
\DeclareTextFontCommand{\textarmenian}{\armenian}

\usepackage{microtype}
\aclfinalcopy

\title{One Model to Pronounce Them All: Multilingual Grapheme-to-Phoneme Conversion With a Transformer Ensemble}

\author{Kaili Vesik$^{1,2}$, Muhammad Abdul-Mageed$^{1,2,3}$,  Miikka Silfverberg$^2$ \\
$^1$ Natural Language Processing Lab\\ 
    $^2$ Department of Linguistics\\
    $^3$ School of Information \\
      The University of British Columbia  \\
     \small{$^1$ \{kaili.vesik,muhammad.mageed,miikka.silfverberg\}@ubc.ca}}

\date{}

\begin{document}
\maketitle
\begin{abstract}
The task of grapheme-to-phoneme (G2P) conversion is important for both speech recognition and synthesis. Similar to other speech and language processing tasks, in a scenario where only small-sized training data are available, learning G2P models is challenging. We describe a simple approach of exploiting model ensembles, based on multilingual Transformers and self-training, to develop a highly effective G2P solution for 15 languages. Our models are developed as part of our participation in the SIGMORPHON 2020 Shared Task 1 focused at G2P. Our best models achieve 14.99 word error rate (WER) and	3.30 phoneme error rate (PER), a sizeable improvement over the shared task competitive baselines.  

\end{abstract}

\section{Introduction}\label{sec:intro}
Speech technologies are becoming increasingly pervasive in our lives. The task of \textit{grapheme-to-phoneme (G2P)} conversion is an important component of both speech recognition and synthesis. In G2P conversion, sequences of graphemes (the symbols used to write words) are mapped to corresponding phonemes (pronunciation symbols, e.g., symbols of the International Phonetic Alphabet). Members of the Special Interest Group on Computational Morphology and Phonology (SIGMORPHON) have proposed a G2P shared task (SIGMORPHON 2020 Shared Task 1)~\footnote{The shared task webpage is accessible at: \url{https://sigmorphon.github.io/sharedtasks/2020/task1}.} involving multiple languages. In this paper, we describe our submissions to the shared task. Organizers provide an overview of the task and submitted systems in~\newcite{Task1} (this volume).

The task was introduced with data from 10 languages, with an additional 5 `surprise' languages released during the task timeline. Our goal was to develop an effective system based on modern deep learning methods as a solution. However, deep learning technologies work best with sufficiently large training data. Hence, a clear challenge we came across is the limited size of the shared task training data for each of the 15 individual languages. To ease this bottleneck, we decided to view the task through a \textit{multilingual machine translation lens} where we build a single model mapping from input to output across all the languages simultaneously. In this, we hypothesized that a multilingual model would allow for shared representations across the various languages that may be more powerful than individual representations of monolingual models. Abundant evidence now exists for approaching machine translation tasks from a multilingual perspective~\citep{johnson-etal-2017-googles,dong-etal-2015-multi,firat-etal-2016-multi}, which inspired our choice. 

In order to make use of unlabeled data, we also explore a straightforward \textit{self-training approach}. In particular, we employ our trained models to convert sequences of multilingual unlabeled graphemes, taken from Wikipedia data, into multilingual phonemes. We then select sequences of phonemes predicted with our models above a certain confidence threshold to augment the shared task training data, thus re-training our models with larger (gold and silver) training data from scratch. Our models are based on the Transformer architecture which exploits effective self-attention. We show that both our multilingual model and the self-trained variation outperform the results of the competitive baseline monolingual models provided by the task organizers. Ultimately, we demonstrate how our simple modeling choices enable us to provide an effective solution to the problem in spite of the low-resource challenge. Intrinsically, our approach also enjoys the simplicity of a single model rather than 15 different models. 

The rest of the paper is organized as follows: Section~\ref{sec:data-eval-bl} is a description of the shared task data, evaluation metrics, and baselines. Section~\ref{sec:models} introduces both our fully supervised, multilingual models (Section~\ref{sec:multilingual}) and self-trained model (Section~\ref{sec:selftraining}). We present our results in Section~\ref{sec:results}. We provide an analysis of results and report on an ablation study in Section~\ref{sec:analysis-ablation}. We overview related work in Section~\ref{sec:rel}, and conclude in Section~\ref{sec:conc}.

\section{Task Data, Evaluation, and Baselines}\label{sec:data-eval-bl}

The data provided by the organizers of the shared task are extracted from Wiktionary~\footnote{\url{https://www.wiktionary.org/}.} using the WikiPron library~\citep{WikiPron}, and consist of 4,050 gold labeled grapheme-phoneme pairs for each of 15 languages, split into a \texttt{training} set (3,600 per language) and a \texttt{development} set (450 per language). The blind \texttt{test} data comprise 450 sources for each language. The data involves languages in the set \textit{\{Adyghe (ady), Armenian (arm), Bulgarian (bul), Dutch (dut), French (fre), Georgian (geo), Modern Greek (gre), Hindi (hin), Hungarian (hun), Icelandic (ice), Japanese hiragana (jpn), Korean (kor), Lithuanian (lit), Romanian (rum), Vietnamese (vie)\}}.~\footnote{We use three-character ISO-639-2 abbreviations as not all of the task languages have ISO-639-1 codes.} This set of languages employ a variety of writing systems: \textit{alphabets} (e.g. French), \textit{alphasyllabary} (e.g. Hindi), and \textit{syllabary} (e.g. Japanese hiragana), thus introducing enough diversity and modelling challenge. Table~\ref{tab:sampledata} shows sample pairs from training data across 5 languages.

\begin{table}[h]
    \centering
    \begin{tabular}{l l l} \hline
    \textbf{Language} & \textbf{Source} & \textbf{Target (IPA)} \\
    \hline
    \multicolumn{2}{l}{\textit{Alphabet:}} &\\
    \multirow{2}{*}{arm} & \textarmenian{ահեղ} & \textscripta~h~\textepsilon~\textinvscr \\
    & \textarmenian{լիարժեք} & l~j~\textscripta~\textfishhookr~\textyogh~\textepsilon~k\textsuperscript{h} \\
    \multirow{2}{*}{fre} & front & f~\textinvscr~\~\textopeno \\
    & v\^etu & v~e~t~y \\
    \hline
    \multicolumn{2}{l}{\textit{Alphasyllabary:}} &\\
    \multirow{2}{*}{hin} & {\dn EdKAvA} & \textsubbridge{d}~\textsci~k\textsuperscript{h}~\textscripta\textlengthmark~\textscriptv~\textscripta\textlengthmark \\ 
    & {\dn hVnA} & \texthth~\textschwa~\textrtailt~n~\textscripta\textlengthmark \\ 
    \multirow{2}{*}{kor} & \begin{CJK}{UTF8}{mj}개벽\end{CJK} & k~\textlowering{e}~b~j~\textsubrhalfring{\textturnv}~k\textcorner \\
     & \begin{CJK}{UTF8}{mj}오빠\end{CJK} & \textlowering{o}~\textsubumlaut{p}~\textsubbar{a} \\
    \hline
    \multicolumn{2}{l}{\textit{Syllabary:}} &\\
    \multirow{2}{*}{jpn} & \begin{CJK}{UTF8}{min}いなり\end{CJK} & i~n~\textsubbar{a}~\textfishhookr\textsuperscript{j}~i \\
    & \begin{CJK}{UTF8}{min}やせん\end{CJK} & j~\textsubbar{a}~s~\textlowering{\~e}~\textsc{n} \\ 
    \hline
    \end{tabular}
    \caption{Sample pairs from training data}
    \label{tab:sampledata}
\end{table}

\textbf{Evaluation.} 
For evaluation, the task organizers use both Word Error Rate (WER) and Phoneme Error Rate (PER). WER is the percentage of words whose predicted transcription does not match the gold transcription; PER is the micro-averaged edit distance between predicted and gold transcriptions. We follow this set-up in evaluating our models on the development data as well, as reported in this paper.

\textbf{Baselines.} 
Organizers provide a number of monolingual baselines. The first is a pair n-gram model encoded as a weighted finite-state transducer (FST), implemented using the OpenGRMtoolkit~\footnote{\url{http://www.opengrm.org/twiki/bin/view/GRM}.}. The second is a bi-LSTM encoder-decoder sequence model implemented using the Fairseq toolkit~\footnote{\url{https://github.com/pytorch/fairseq}.}. The third is a Transformer model also implemented using the Fairseq toolkit. Organizer-provided shared task baselines are shown in Table~\ref{tab:baselines} as WER and PER averages over the 15 languages. We now introduce our models.

\begin{table}[h]
    \centering
    \begin{tabular}{l c rr c rr c rr}
    \toprule
             && \multicolumn{2}{c}{\textbf{Avg over 15 langs}}\\ 
    \cmidrule{3-4}
    Model    &&   WER &   PER   \\
    \midrule
    FST && 22.00 &  4.92  \\
    Bi-LSTM && 16.84 &  3.99  \\
    Transformer && 17.51 &  4.30   \\
    \bottomrule
    \end{tabular}
    \caption{Baseline performance as avg. WER and PER over the 15 languages as provided by task organizers. Baselines exploit monolingual models.}
    \label{tab:baselines}
\end{table}

\section{Models}\label{sec:models}
As explained, our models are based on Transformers and we offer two primary types of models, depending on how we supervise each. We first introduce \textit{fully supervised} multilingual models, then we introduce our \textit{semi-supervised} models (also multilingual). Our semi-supervised models follow a self-training set up. We now explain each of these models.

\subsection{Supervised, Multilingual Models}\label{sec:multilingual}
We use a multilingual approach where we train a single model on data from all 15 languages. For this purpose, we prepend a token comprising a language code (e.g. \texttt{fre}) to each grapheme sequence source. For our implementation, we use the PyTorch Transformer architecture in the OpenNMT Neural Machine Translation Toolkit~\cite{opennmt}. We set the model hyper-parameters as shown in Table~\ref{tab:parameters}, which follows those adopted by~\citet{vaswani2017attention}.
\begin{table}[h]
    \centering
    \begin{tabular}{l l}
        \hline
    \textbf{Hyper-Parameter} & \textbf{Value} \\ \hline
    Number of layers & 6 \\
    Hidden state size & 512 \\
    Word embedding size & 512 \\
    Hidden feed-forward size & 2,048 \\
    Number of self-attention heads & 8 \\
    Optimizer & Adam \\
    Dropout probability & 0.1 \\
    Number of training steps & 200K \\ \hline
    \end{tabular}
    \caption{Multilingual Transformer hyper-parameters.}
    \label{tab:parameters}
\end{table}

We train the model with 3 different random seeds, and at inference we employ an ensemble consisting of the models from 4 training checkpoints (at 50k, 100k, 150k, and 200k steps) for each of the 3 models generated by the random seeds. We note that OpenNMT averages individual models' prediction distributions, which is how we deploy our ensemble. We use beam search with the OpenNMT default beam width of 5.~\footnote{We also experimented with beam size 10, but did not obtain improvements on the development set.} 

\subsection{Self-Trained Model}\label{sec:selftraining}

\subsubsection{Wikipedia Data Augmentation}\label{sec:wikidata}
One of the models we submitted to the task employs a self-training approach, as a way to augment training data. The additional data is sourced from Wikipedia articles from 12 of the 15 languages (excluding Adyghe, Japanese, and Vietnamese)~\footnote{We note that there is no Adyghe Wikipedia. Also, the Japenese Wikipedia is not strictly in Hiragana and so we exclude it. By mistake, we did not include Vietnamese either. Clearly, we average results from the self-training models only on the languages for which we augment the data.}. We download the Wikipedia dumps from the Wikimedia website~\footnote{\url{https://dumps.wikimedia.org/}.} and use an off-the-shelf tool~\footnote{\url{https://github.com/attardi/wikiextractor}} for extracting text. Further pre-processing involved removing any remaining XML markup, discarding leading and trailing punctuation and numerals for each word, and ignoring any words with remaining word-internal punctuation or numerals.\\
Due to time constraints, only one million words from each language were used, and from those only unique entries were submitted to the model for translation and subsequent evaluation as potential candidates for augmenting training data. Table~\ref{tab:wikisources} summarizes the size of the Wikipedia data used for each available language. Selection methods and thresholds are discussed in Section~\ref{sec:procedure}.

\begin{table}[h]
    \centering
    \begin{tabular}{l l l}
    \textbf{Language} & \textbf{Translated} & \textbf{Selected} \\
    \hline
    arm & 9,947 & 4,723  \\
    bul & 9,999 & 3,197  \\
    dut & 2,275 & 860  \\
    fre & 9,985 & 2,888  \\
    geo & 5,038 & 3,043  \\
    gre & 9,949 & 3,419  \\
    hin & 1,450 & 727  \\
    hun & 10,000 & 3,444  \\
    ice & 9,839 & 3,719  \\
    kor & 4,282 & 2,681  \\
    lit & 7,033 & 3,615  \\
    rum & 9,785 & 3,102  \\
    \hline
    \textbf{Total} & 89,582 & 35,418 \\ \hline
    \end{tabular}
    \caption{Number of Wikipedia words translated vs. number of words selected for self-training. 
    }
    \label{tab:wikisources}
\end{table}

\subsubsection{Procedure}\label{sec:procedure}
As explained, self-training data is drawn from the translations of Wikipedia text in 12 languages as predicted by an ensemble model. In order to select pairs to augment the training set, we first calculate the mean per-class softmax value in the development set (which we find to be at $0.11$).~\footnote{As is known, the softmax function produces a probability distribution over the classes.} Comparatively, the average per-class softmax value for the predicted Wikipedia targets for each language ranges from $0.12$ to $0.30$. Based on this analysis, we select only those Wikipedia pairs whose predicted targets have a probability greater than $0.2$.~\footnote{There could be different ways to select predicted data for augmentation. For example, one can arbitrarily choose the top \textit{n\%} most confidently predicted points (with \textit{n} being a hyper-parameter).} The selected data are combined with the original (i.e., from official task) training set and the models are re-trained using the same hyper-parameters as the fully-supervised setting.

\section{Results}\label{sec:results}

\begin{table}[h]
    \centering
    \begin{tabular}{l c rr c rr}
    \toprule
             && \multicolumn{2}{c}{\textbf{Multilingual}}
             && \multicolumn{2}{c}{\textbf{Self-trained}}\\ 
    \cmidrule{3-4} \cmidrule{6-7}
    Lang    &&   WER &   PER &&           WER  &           PER  \\
    \midrule
    ady && 25.56 &  6.40 &&         25.11  &          6.47  \\
    arm && 16.67 &  3.37 &&         16.89  &          3.37  \\
    bul && 28.44 &  7.30 &&         27.33  &          7.12  \\
    dut && 16.00 &  2.84 &&         15.33  &          2.84 \\
    fre && 8.22  &  1.96 &&          8.44  &          1.92 \\
    geo && 24.44 &  4.92 &&         26.22  &          5.22  \\
    gre && 15.11 &  2.72 &&         16.22  &          3.00  \\
    hin && 6.44  &  1.66 &&          6.89  &          1.89 \\
    hun &&  2.89 &  0.54 &&          3.56  &          0.66  \\
    ice &&  9.56 &  1.88 &&         10.89  &          2.23  \\
    jpn &&  7.33 &  2.18 &&          7.11  &          2.11  \\
    kor && 24.22 &  6.54 &&         26.00  &          6.50  \\
    lit && 20.00 &  4.11 &&         21.11  &          3.96 \\
    rum && 12.00 &  2.94 &&         11.78  &          2.97 \\
    vie &&  5.56 &  1.77 &&          5.56  &          1.91  \\
    \midrule
    avg && 14.83 &  3.41 &&         15.23  &          3.48 \\
    \bottomrule
    \end{tabular}
    \caption{Development set results for \textit{fully-supervised multilingual} and \textit{self-trained multilingual} models.}
    \label{tab:devresults}
\end{table}

\begin{table}[h]
    \centering
    \begin{tabular}{l c rr c rr}
    \toprule
             && \multicolumn{2}{c}{\textbf{Multilingual}}
             && \multicolumn{2}{c}{\textbf{Self-trained}}\\ 
    \cmidrule{3-4} \cmidrule{6-7}
    Lang    &&   WER &   PER &&           WER  &           PER  \\
    \midrule
    ady && 28.44 &	6.46 &&     	29.11  &          6.46  \\
    arm && 13.11 &	2.98 &&     	12.89  &     	  3.07  \\
    bul && 27.11 &	5.91 &&     	30.89  &    	  6.92  \\
    dut && 15.78 &	2.98 &&     	16.89  &    	  3.07 \\
    fre && 5.33	 &  1.24 &&     	5.78   &    	  1.36 \\
    geo && 26.00 &	5.25 &&     	26.67  &    	  5.23  \\
    gre && 16.67 &	2.68 &&     	15.78  &    	  2.60 \\
    hin && 6.44	 &  1.58 &&      	 6.67  &    	  1.66 \\
    hun && 4.67	 &  1.05 &&     	 4.22  &    	  0.98  \\
    ice &&  9.56 &	2.11 &&     	 9.11  &          1.83  \\
    jpn &&  6.00 &	1.44 &&     	 6.00  &	      1.40  \\
    kor && 32.22 &	8.54 &&     	32.44  &          8.86 \\
    lit && 19.33 &	3.63 &&	        20.00  &	      3.68 \\
    rum && 9.33	 &  1.96 &&	        10.44  &          2.23 \\
    vie &&  4.89 &  1.66 &&	         4.00  &    	  1.28  \\
    \midrule
    avg && 14.99 &  3.30 &&         15.39  &          3.37 \\
    \bottomrule
    \end{tabular}
    \caption{Blind test set results for \textit{fully-supervised multilingual} and \textit{self-trained multilingual} models.}
    \label{tab:testresults}
\end{table}

\begin{table*}
    \centering
    \begin{tabular}{l l l l} \hline
    \textbf{Lang} & \textbf{Source} & \textbf{Target} & \textbf{Prediction} \\
    \hline
    \multirow{2}{*}{arm} & \textarmenian{զուգարան} & z~u~k\textsuperscript{h}~\textscripta~\textfishhookr~\textscripta~n	& z~u~\textscriptg~\textscripta~\textfishhookr~\textscripta~n \\
    & \textarmenian{անխնա} & \textscripta~\textipa{\ng}~\textchi~\textschwa~n~\textscripta & \textscripta~\textipa{\ng}~\textchi~n~\textscripta  \\
    \multirow{2}{*}{fre} & full & f~u~l & f~y~l \\
    & proulx & p~\textinvscr~u & p~\textinvscr~u~l \\ 
    \multirow{2}{*}{hin} & {\dn D\306wy} & \textsubbridge{d}\textsuperscript{\texthth}~\textschwa~n~j~\textschwa & \textsubbridge{d}\textsuperscript{\texthth}~\textschwa~n~j \\ 
    & {\dn m\?hrbAnF} & m~\textepsilon\textlengthmark\textsuperscript{\texthth}~\textfishhookr~b~\textscripta\textlengthmark~n~i\textlengthmark & m~e\textlengthmark~\texthth~\textschwa~\textfishhookr~b~\textscripta\textlengthmark~n~i\textlengthmark \\ 
    \multirow{2}{*}{jpn} & \begin{CJK}{UTF8}{min}こたま\end{CJK} & k~\textlowering{o}~d~\textsubbar{a}~m~\textsubbar{a} & k~\textlowering{o}~d~\textsubbar{a}~m~\textsubbar{a} \\
    & \begin{CJK}{UTF8}{min}ひぞう\end{CJK} & \c{c}~i~z~o\textlengthmark & \c{c}~i~z~\textlowering{o}\textlengthmark \\ 
    \multirow{2}{*}{rum} & ceri & t~\textesh~e~r\textsuperscript{j} & \texttoptiebar{\textteshlig}~e~r\textsuperscript{j} \\
    & iubeau & j~u~b\textsuperscript{j}~\ae~u & j~u~b~\textsubarch{e}~a~w \\ 
    \hline
    \end{tabular}
    \caption{Sample prediction errors from development data.}
    \label{tab:errors}
\end{table*}

Both models demonstrate lower word error rates (WER) and phoneme error rates (PER), averaged across languages, than the baseline monolingual models provided by the task organizers (see Table~\ref{tab:baselines} in Section~\ref{sec:data-eval-bl}). Error rates per language are shown in Table~\ref{tab:devresults} for the development set and Table~\ref{tab:testresults} for the blind test set (results published by organizers). Table~\ref{tab:errors} shows examples of prediction errors, which demonstrate some of the typical minor errors in phenomena such as voicing (e.g. k vs. \textscriptg), epenthesis and elision (e.g. p~\textinvscr~u vs. p~\textinvscr~u~l), and coarticulation (e.g. b\textsuperscript{j} vs. b).

On average, the fully-supervised models performed slightly better than the self-trained model. We expected that the self-trained model would see (at least slightly) better performance than the fully supervised; however, due to time constraints, we were not able to augment the training data to such a degree that this hypothesized improvement would be tangible. We leave it as a question for the future whether, and if so to what extent, self-training can improve our models. We now provide an analysis of our findings and report on an ablation study under a number of settings.

\section{Analysis \& Ablation Study}\label{sec:analysis-ablation}

We suspected that languages with shared writing systems (in our multilingual models) would benefit from the shared representation and hence see better results, posing a challenges to those languages with unique orthography (i.e., orthography not shared by o=any of the other languages considered). However, our results do not support this hypothesis; there did not appear to be a significant correlation between writing system and results on G2P conversion. For example, a total of 7 of the languages (i.e., dut, fre, hun, ice, lit, rum, vie) use the Roman alphabet, but the WERs for these languages cover a reasonably wide range (from first- to eleventh-best) of the results. It is worth noting, however, that the two languages that use the Cyrillic alphabet (ady, bul) were the two worst-performing languages of the set.

Both prior and subsequent to the task deadline, we performed several ablations in order to assess the effectiveness of our approach. First, we compare results based on single models vs. those based on the ensemble. Table~\ref{tab:ckptresults} shows the error rates of development set translation by the four training checkpoints used in the ensemble, in this case trained with the default (random) seed. Given that each of these results is poorer than our ensemble results for the multilingual model (WER 14.83 / PER 3.41), it is clear that the ensemble approach is superior. Clearly, the ensemble has the advantage of exploiting multiple predictions for each word. This does result in reduced error rates as compared to individual models. 

\begin{table}[h]
    \centering
    \begin{tabular}{l c rr }
    \toprule
             && \multicolumn{2}{c}{\textbf{Avg over 15 langs}}\\ 
    \cmidrule{3-4}
    Checkpoint    &&   WER &   PER   \\
    \midrule
    50k of 200k steps && 16.70 &  3.93  \\
    100k of 200k steps && 16.04 &  3.69  \\
    150k of 200k steps && 16.25 &  3.78   \\
    200k of 200k steps && 15.73 &  3.65 \\ \hline 
    Ensemble && 14.83  & 3.41 \\
    \bottomrule
    \end{tabular}
    \caption{Development set results for individual models vs. our ensemble}
    \label{tab:ckptresults}
\end{table}

We also compare our multilingual model's error rates on a given language to those acquired by the respective monolingual models. We note that each of the monolingual models is otherwise initialized with the same parameters as the multilingual model described in Section~\ref{sec:multilingual}. Results for the 15 monolingual models are shown in Table~\ref{tab:monoresults}. The average WER across all languages is almost twice as big as that of our multilingual model (whether individual or ensemble), and the per-language results are worse across the board as well. The monolingual Georgian WER (25.33) was the only result to approach its multilingual counterpart (24.44). \textbf{\textit{Our multilingual approach is clearly a significant improvement over otherwise equivalent monolingually-trained models}}.

\begin{table}[h]
    \centering
    \begin{tabular}{l c rr}
    \toprule
             && \multicolumn{2}{c}{\textbf{Monolingual}}\\ 
    \cmidrule{3-4} 
    Lang    &&   WER &   PER\\
    \midrule
    ady && 33.56 &  9.31  \\
    arm && 24.00 &  5.65 \\
    bul && 41.33 &  12.07 \\
    dut && 30.89 &  7.73 \\
    fre && 34.89 &  12.69 \\ 
    geo && 25.33 &  5.19 \\
    gre && 24.00 &  5.13 \\
    hin && 22.67 &  6.76 \\
    hun && 20.89 &  5.30 \\
    ice && 30.22 &  11.12 \\
    jpn && 11.78 &  3.73 \\
    kor && 30.67 &  9.17 \\
    lit && 26.00 &  7.75 \\
    rum && 20.00 &  5.52 \\
    vie && 32.00 &  13.75 \\
    \midrule
    avg && 27.22 &  8.06 \\
    \bottomrule
    \end{tabular}
    \caption{Development set results for monolingual models.}
    \label{tab:monoresults}
\end{table}

\section{Related Work}\label{sec:rel}

Various data-driven models have been successfully applied to G2P conversion. In terms of English conversion, \newcite{bisani2008joint} use co-segmentation and joint sequence models for early data-driven G2P. \newcite{novak2016phonetisaurus} employ a joint multigram approach to generate weighted finite-state transducers for G2P. Recently, neural sequence-to-sequence models based on CNN and RNN architectures have been proposed for the G2P task delivering superior results compared to earlier non-neural approaches \citep{chae2018convolutional,yolchuyeva2019grapheme}. Similar to our approach, \newcite{Yolchuyeva_2019}  use transformers \citep{vaswani2017attention} to perform English G2P conversion.

Multilingual training is a crucial component in our system. Our approach is closely related to multilingual neural machine translation \cite{johnson2017google}, where a single model is trained to translate between multiple source and target languages. Others have also explored multilingual approaches to G2P.   \newcite{deri2016grapheme} use multilingual G2P conversion for the purpose of adapting models from high-resource languages to train  weighted finite-state transducers for related low-resource languages.  \newcite{ni2018multilingual} experiment with multilingual training for deep learning models. They use pretrained character embeddings with LSTM encoder-decoders in order to train multilingual G2P models for Chinese, Japanese, Korean and Thai. In contrast to \newcite{ni2018multilingual}, we inspect multilingual training in the context of transformer models.

For our second model, whose training data is augmented from Wikipedia, we use a self-taining method. \newcite{sun2019token} investigate self-training together with ensemble distillation for English G2P conversion, using transformer models. Their setting resembles ours: A teacher model is first trained using a gold standard labeled G2P training set. The teacher model is then used to label additional grapheme data, producing a silver standard training set. Subsequently, a model ensemble is trained on the combination of the gold and silver data. \newcite{sun2019token} train on nearly 200k gold standard examples and 2M silver standard examples and report small improvements. In contrast, we do not observe improvements from self-training. This might be a consequence of the small size of the shared task datasets and our silver standard Wikipedia data.

\section{Conclusion}\label{sec:conc}

We introduced a multilingual approach to G2P conversion, exploiting Transformers in a fully supervised multilingual setting. Strikingly, our choice to model all languages in a shared, multilingual space reduces error rates (in WER and PER) by almost one half. We also showed how an ensemble of individually-trained multilingual Transformers, is an improvement over non-ensemble models. We also leveraged multilingual Wikipedia data via a self-training strategy, though due to time constraints we were not able to incorporate enough silver labeled data into training to see the results we had hoped for\footnote{Training on all available Wikipedia data is in progress at the time of this paper's submission}. Nevertheless, the multilingual models successfully surpassed all organizer-provided baselines on the task and compared favorably to several other submitted models. Our future work includes scaling up our self-training with larger Wikipedia data and choosing fully-trained models (e.g., in our case ones at 200K steps) to include in the ensemble.

\section*{Acknowledgements}\label{sec:ack}
We acknowledge the support of the Natural Sciences and Engineering Research Council of Canada (NSERC), the Social Sciences Research Council of Canada (SSHRC), and Compute Canada (\url{www.computecanada.ca}).\\

\bibliography{sigmorphon2020_ubcnlp}

\begin{thebibliography}{16}
\expandafter\ifx\csname natexlab\endcsname\relax\def\natexlab#1{#1}\fi

\bibitem[{Bisani and Ney(2008)}]{bisani2008joint}
Maximilian Bisani and Hermann Ney. 2008.
\newblock Joint-sequence models for grapheme-to-phoneme conversion.
\newblock \emph{Speech communication}, 50(5):434--451.

\bibitem[{Chae et~al.(2018)Chae, Park, Bang, Suh, Park, Kim, and
  Park}]{chae2018convolutional}
Moon-jung Chae, Kyubyong Park, Jinhyun Bang, Soobin Suh, Jonghyuk Park, Namju
  Kim, and Longhun Park. 2018.
\newblock Convolutional sequence to sequence model with non-sequential greedy
  decoding for grapheme to phoneme conversion.
\newblock In \emph{2018 IEEE International Conference on Acoustics, Speech and
  Signal Processing (ICASSP)}, pages 2486--2490. IEEE.

\bibitem[{Deri and Knight(2016)}]{deri2016grapheme}
Aliya Deri and Kevin Knight. 2016.
\newblock Grapheme-to-phoneme models for (almost) any language.
\newblock In \emph{Proceedings of the 54th Annual Meeting of the Association
  for Computational Linguistics (Volume 1: Long Papers)}, pages 399--408.

\bibitem[{Dong et~al.(2015)Dong, Wu, He, Yu, and Wang}]{dong-etal-2015-multi}
Daxiang Dong, Hua Wu, Wei He, Dianhai Yu, and Haifeng Wang. 2015.
\newblock \href {https://doi.org/10.3115/v1/P15-1166} {Multi-task learning for
  multiple language translation}.
\newblock In \emph{Proceedings of the 53rd Annual Meeting of the Association
  for Computational Linguistics and the 7th International Joint Conference on
  Natural Language Processing (Volume 1: Long Papers)}, pages 1723--1732,
  Beijing, China. Association for Computational Linguistics.

\bibitem[{Firat et~al.(2016)Firat, Cho, and Bengio}]{firat-etal-2016-multi}
Orhan Firat, Kyunghyun Cho, and Yoshua Bengio. 2016.
\newblock \href {https://doi.org/10.18653/v1/N16-1101} {Multi-way, multilingual
  neural machine translation with a shared attention mechanism}.
\newblock In \emph{Proceedings of the 2016 Conference of the North {A}merican
  Chapter of the Association for Computational Linguistics: Human Language
  Technologies}, pages 866--875, San Diego, California. Association for
  Computational Linguistics.

\bibitem[{Gorman et~al.(2020)Gorman, Ashby, Goyzueta, McCarthy, Wu, and
  You}]{Task1}
Kyle Gorman, Lucas~F.E. Ashby, Aaron Goyzueta, Arya~D. McCarthy, Shijie Wu, and
  Daniel You. 2020.
\newblock The {SIGMORPHON} 2020 shared task on multilingual grapheme-to-phoneme
  conversion.
\newblock In \emph{Proceedings of the 17th SIGMORPHON Workshop on Computational
  Research in Phonetics, Phonology, and Morphology}.

\bibitem[{Johnson et~al.(2017{\natexlab{a}})Johnson, Schuster, Le, Krikun, Wu,
  Chen, Thorat, Vi{\'e}gas, Wattenberg, Corrado, Hughes, and
  Dean}]{johnson-etal-2017-googles}
Melvin Johnson, Mike Schuster, Quoc~V. Le, Maxim Krikun, Yonghui Wu, Zhifeng
  Chen, Nikhil Thorat, Fernanda Vi{\'e}gas, Martin Wattenberg, Greg Corrado,
  Macduff Hughes, and Jeffrey Dean. 2017{\natexlab{a}}.
\newblock \href {https://doi.org/10.1162/tacl_a_00065} {{G}oogle{'}s
  multilingual neural machine translation system: Enabling zero-shot
  translation}.
\newblock \emph{Transactions of the Association for Computational Linguistics},
  5:339--351.

\bibitem[{Johnson et~al.(2017{\natexlab{b}})Johnson, Schuster, Le, Krikun, Wu,
  Chen, Thorat, Vi{\'e}gas, Wattenberg, Corrado et~al.}]{johnson2017google}
Melvin Johnson, Mike Schuster, Quoc~V Le, Maxim Krikun, Yonghui Wu, Zhifeng
  Chen, Nikhil Thorat, Fernanda Vi{\'e}gas, Martin Wattenberg, Greg Corrado,
  et~al. 2017{\natexlab{b}}.
\newblock Google’s multilingual neural machine translation system: Enabling
  zero-shot translation.
\newblock \emph{Transactions of the Association for Computational Linguistics},
  5:339--351.

\bibitem[{Klein et~al.(2017)Klein, Kim, Deng, Senellart, and Rush}]{opennmt}
Guillaume Klein, Yoon Kim, Yuntian Deng, Jean Senellart, and Alexander~M. Rush.
  2017.
\newblock \href {https://doi.org/10.18653/v1/P17-4012} {Open{NMT}: Open-source
  toolkit for neural machine translation}.
\newblock In \emph{Proc. ACL}.

\bibitem[{Lee et~al.(2020)Lee, Ashby, Garza, Lee-Sikka, Miller, Wong, McCarthy,
  and Gorman}]{WikiPron}
Jackson~L. Lee, Lucas~F.E. Ashby, M.~Elizabeth Garza, Yeonju Lee-Sikka, Sean
  Miller, Alan Wong, Arya~D. McCarthy, and Kyle Gorman. 2020.
\newblock \href {https://www.aclweb.org/anthology/2020.lrec-1.521} {Massively
  multilingual pronunciation mining with {WikiPron}}.
\newblock In \emph{Proceedings of the 12th Language Resources and Evaluation
  Conference}, pages 4216--4221, Marseille.

\bibitem[{Ni et~al.(2018)Ni, Shiga, and Kawai}]{ni2018multilingual}
Jinfu Ni, Yoshinori Shiga, and Hisashi Kawai. 2018.
\newblock Multilingual grapheme-to-phoneme conversion with global character
  vectors.
\newblock In \emph{Interspeech}, pages 2823--2827.

\bibitem[{Novak et~al.(2016)Novak, Minematsu, and
  Hirose}]{novak2016phonetisaurus}
Josef~Robert Novak, Nobuaki Minematsu, and Keikichi Hirose. 2016.
\newblock Phonetisaurus: Exploring grapheme-to-phoneme conversion with joint
  n-gram models in the wfst framework.
\newblock \emph{Natural Language Engineering}, 22(6):907--938.

\bibitem[{Sun et~al.(2019)Sun, Tan, Gan, Liu, Zhao, Qin, and
  Liu}]{sun2019token}
Hao Sun, Xu~Tan, Jun-Wei Gan, Hongzhi Liu, Sheng Zhao, Tao Qin, and Tie-Yan
  Liu. 2019.
\newblock Token-level ensemble distillation for grapheme-to-phoneme conversion.
\newblock \emph{arXiv preprint arXiv:1904.03446}.

\bibitem[{Vaswani et~al.(2017)Vaswani, Shazeer, Parmar, Uszkoreit, Jones,
  Gomez, Kaiser, and Polosukhin}]{vaswani2017attention}
Ashish Vaswani, Noam Shazeer, Niki Parmar, Jakob Uszkoreit, Llion Jones,
  Aidan~N. Gomez, Lukasz Kaiser, and Illia Polosukhin. 2017.
\newblock \href {http://arxiv.org/abs/1706.03762} {Attention is all you need}.

\bibitem[{Yolchuyeva et~al.(2019{\natexlab{a}})Yolchuyeva, N{\'e}meth, and
  Gyires-T{\'o}th}]{yolchuyeva2019grapheme}
Sevinj Yolchuyeva, G{\'e}za N{\'e}meth, and B{\'a}lint Gyires-T{\'o}th.
  2019{\natexlab{a}}.
\newblock Grapheme-to-phoneme conversion with convolutional neural networks.
\newblock \emph{Applied Sciences}, 9(6):1143.

\bibitem[{Yolchuyeva et~al.(2019{\natexlab{b}})Yolchuyeva, Németh, and
  Gyires-Tóth}]{Yolchuyeva_2019}
Sevinj Yolchuyeva, Géza Németh, and Bálint Gyires-Tóth. 2019{\natexlab{b}}.
\newblock \href {https://doi.org/10.21437/interspeech.2019-1954} {Transformer
  based grapheme-to-phoneme conversion}.
\newblock \emph{Interspeech 2019}.

\end{thebibliography}
\bibliographystyle{acl_natbib}

\end{document}